\documentclass[journal,12pt, onecolumn]{IEEEtran}

\usepackage{cite}
\usepackage{amsmath,amssymb,amsfonts}
\usepackage{algorithmic}
\usepackage{graphicx}
\usepackage{textcomp}
\usepackage{xcolor}
\usepackage{subfigure}
\usepackage{amsthm}
\usepackage{mathrsfs}

\usepackage{pdfpages}
\usepackage{colortbl}
\definecolor{mygray}{gray}{0.85}
\usepackage{array}
\usepackage{caption}
\usepackage{setspace}
\usepackage[linesnumbered,boxed,ruled,commentsnumbered]{algorithm2e}

\usepackage[ruled,linesnumbered]{algorithm2e}

\usepackage{amsmath,bm}
\usepackage{cite}
\usepackage{amsmath,amssymb,amsfonts}
\usepackage{algorithmic}
\usepackage{amsthm}
\usepackage{graphicx}
\usepackage{textcomp}
\usepackage{xcolor}

\begin{document}
\begin{spacing}{1.5}
%
\title{
\huge
Lessons Learned from Accident of Autonomous Vehicle Testing: An Edge Learning-aided Offloading Framework
}

\author{Bo~Yang,~\IEEEmembership{Member,~IEEE,} Xuelin~Cao,
 Xiangfang~Li,~\IEEEmembership{Senior Member,~IEEE,} \\
  Chau Yuen,~\IEEEmembership{Senior Member,~IEEE}, and~Lijun Qian,~\IEEEmembership{Senior Member,~IEEE}
\thanks{Bo Yang was with the Department of Electrical and Computer Engineering and CREDIT Center, Prairie View A$\&$M University, Texas A$\&$M University System, Prairie View, TX 77446, USA. He is currently with the Engineering Product Development Pillar, Singapore University of Technology and Design, Singapore, 487372. 
}
\thanks{Xiangfang Li and Lijun Qian are with the Department of Electrical and Computer Engineering and CREDIT Center, Prairie View A$\&$M University, Texas A$\&$M University System, Prairie View, TX 77446, USA. 
}

\thanks{Xuelin Cao and Chau Yuen are with the Engineering Product Development Pillar, Singapore University of Technology and Design, Singapore, 487372. 
}

}


\maketitle

\begin{abstract}
 This letter proposes an edge learning-based offloading framework for autonomous driving, where the deep learning tasks can be offloaded to the edge server to improve the inference accuracy while meeting the latency constraint. Since the delay and the inference accuracy are incurred by wireless communications and computing, an optimization problem is formulated to maximize the inference accuracy subject to the offloading probability, the pre-braking probability, and data quality. Simulations demonstrate the superiority of the proposed offloading framework.
\end{abstract}


\IEEEpeerreviewmaketitle

\section{Introduction and Motivation}
\begin{figure}[t]
  \captionsetup{font={footnotesize }}
\centerline{ \includegraphics[width=4.64in, height=1.65in]{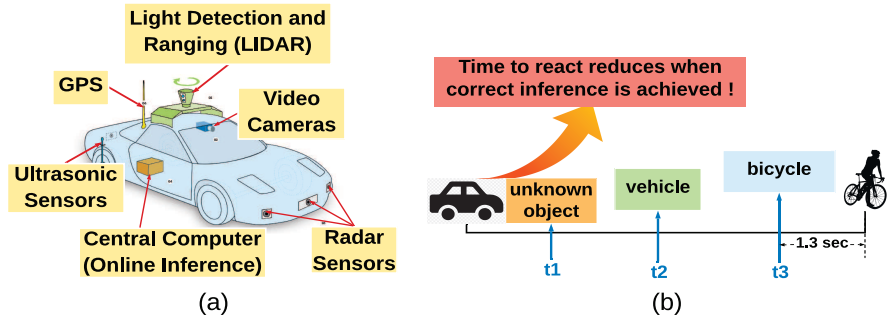}}
\caption{(a) Sensors on a typical self-driving vehicle; (b) Sequence of inference events during the Uber accident}
\label{scene} 
\end{figure}

\IEEEPARstart{A}{long} with the rapid advancement of the artificial intelligence (AI) in industrial automation applications, the empowering deep learning techniques have been extensively used in providing intelligence in autonomous driving to safely navigate their environment. For example, 
fully autonomous driving (level-$5$) without the requirement of human intervention performs all driving functions by analyzing the collected sensing data and reacting accordingly~\cite{MCCA}. However, serious security risks are gradually discovered, which are largely due to the misjudgment inferred by the trained deep learning (DL) model~\cite{Vehicle01}. 
According to the accident report supplied by the National Transportation Safety Board (NTSB)~\cite{NTSB}, an Uber self-driving test vehicle based on a modified 2017 Volvo XC90 struck a pedestrian who walked a bicycle across the road in Arizona at about 9:58 p.m., March 18, 2018. Abide by the basic physical ecosystem of an autonomous vehicle, the Uber self-driving test vehicle involved in the accident was mainly equipped with forward- and side-facing cameras, radars, light detection and ranging (LIDAR), navigation sensors and a computing and data storage unit performing online inference, as illustrated in Fig.~\ref{scene}(a). According to data obtained from the self-driving system in the test vehicle, the pre-trained DL model mistakenly classified the pedestrian as an unknown object first, then as a vehicle, and finally, as a bicycle with varying expectations of the future travel path, as illustrated in Fig.~\ref{scene}(b). Due to the constant erroneous inference results given by the pre-trained DL model, the time cost to achieve the correct inference is increased. In the accident, only about $1.3$ seconds were left for the vehicle to brake before the impact, which is far from enough\footnote{In the Uber accident, the vehicle was traveling at about $43$ mph before the brake. If the vehicle traveled within a low range of speed or the vehicle slows down in advance, the accident may be avoided.}, as shown in Fig.~\ref{scene}(b).
According to the crash event videos captured by the cameras of the Uber self-driving system, the pedestrian was dressed in dark clothing and that the bicycle did not have any side reflectors. Although the bicycle had front and rear reflectors and a forward headlamp, all were facing in directions perpendicular to the path of the oncoming vehicle. Besides, the pedestrian crossed in a section of roadway not directly illuminated by the roadway lighting. These factors directly lead to the quality degradation of the images captured by the vehicle's cameras. As a result, the inference accuracy could be severely degraded, which is generally proportional to the input data quality.

Motivated by the above issues, this letter studies on the vision-based vehicle detection problem and proposes an edge learning-based offloading framework, in which the DL tasks (i.e., objects identification) can be offloaded to a base station (BS) integrated with a MEC server (MES) to achieve higher inference accuracy.
The majority of existing literature on offloading in the vehicle networks aims to upload computation-intensive tasks to the edge servers to improve computing efficiency~\cite{MEC_vehicle}.  
On the other hand, DL techniques at the edge may help deeply dig up the inherent characteristics of the collected big data from heterogeneous sensors and make more reasonable decisions in the vehicle networks. 
However, this has not been fully investigated. Among the few related contributions, Li \textit{et al.}~\cite{DL01} introduced a convolutional neural network (CNN) to predict the road traffic situation and then a proactive load balancing approach was proposed enabling cooperation among mobile edge servers. Liu \textit{et al.}~\cite{DL02} investigated the feature of a Rayleigh fading channel and proposed to train a long short-term memory (LSTM) model to predict the future channel parameters. Additionally, Cheng \textit{et al.}~\cite{DL03} studied a case where two classical supervise machine learning methods were used to detect the Non-Line-of-Sight (NLoS) conditions by learning the V2V measurement data. Nevertheless, none of the above solutions consider the impact of collected data quality to the inference accuracy of the trained DL model, which, however, is the critical reason for the Uber accident. 

 The specific contributions of the letter are summarized as follows. Firstly, we propose an edge learning-aided offloading framework for autonomous driving, in which the self-driving vehicle must decide whether to offload the tasks to minimize the inference error, while meeting the constraint that enough time is reserved to brake the vehicle to avoid the accident. Secondly, since the delay and inference accuracy are incurred by wireless communications and computing, an optimization problem is formulated to minimize the inference error subject to data quality and the probability of offloading and pre-braking. Finally, simulations with practical configurations demonstrate the advantages of the offloading framework.

\theoremstyle{Observation}
\newtheorem{observation}{\textit{Observation}}
\theoremstyle{Lemma} 
\newtheorem{lemma}{{\textit{Lemma}}}

\section{System Model and Problem Formulation}
\begin{figure}[t]
  \captionsetup{font={footnotesize }}
\centerline{ \includegraphics[width=4.52in, height=1.89in]{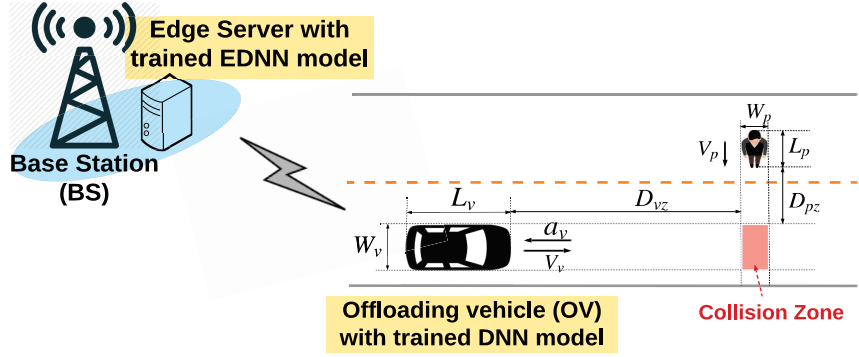}}
\caption{Pre-crash configuration model where the offloading vehicle (OV) is going straight and pedestrian is crossing the road }
\label{v2p_model}
\end{figure}

A pre-crash configuration model is considered in Fig.~\ref{v2p_model}, where a vehicle is going straight and approaching the path of a pedestrian who is crossing the road. In this model, crash countermeasures must account for the positions, velocities, and avoidance maneuvers of both the pedestrian and the vehicle. It is assumed that the pedestrian does not make any avoidance maneuver, so the pedestrian will cross the road at a constant speed. In this case, the vehicle executes the recommended avoidance maneuver (i.e., brake) as soon as it identifies the pedestrian ahead. Furthermore, in this letter, a vehicular network with $M$ offloading vehicles (denoted as $OV_m, \ m\in \mathcal{M}$, and ${\mathcal{M}} =\{1,2,...,M\}$ is the offloading vehicles set) requiring to offload their tasks to the BS via vehicle-to-infrastructure (V2I) connection is considered, as shown in Fig.~\ref{v2p_model}. 

\theoremstyle{Assumption} 
\newtheorem{assumption}{{\textit{Assumption}}}
\begin{assumption}
\label{A1}
\rm Compared to the normal deep neural network (DNN) models such as Inception and MobileNet~\cite{imagenet}, the enhanced deeper neural network (EDNN) models (e.g., ResNet and ResNeXt~\cite{imagenet}) are more complicated, and they usually require much more computing capability including computation resources and storage space. Since the offloading vehicle is generally inferior to the BS in terms of power supply and computing capability, it is difficult for the vehicles to bear the EDNN model. Thus, it is reasonable to assume that an offline trained DNN model is deployed at vehicles while a more powerful EDNN model is deployed at the BS.
\end{assumption}

To characterize the impact of the data quality on the inference error, the deep learning task is defined as follows.

\theoremstyle{Definition} 
\newtheorem{definition}{{\textit{Definition}}}
\begin{definition}
\label{D2}
 \textbf{Deep Learning Task (DLT).} \rm In this letter, the DLT is defined as a computation task processed by a trained DL model. Given a trained DL model (${\cal D}$) and input data with a certain quality ($\cal Q$), the inference error rate can be presented as $\epsilon \!=\! g(\mathcal {Q,D})$, where $g(\cdot)$ performs the mapping function. 
\end{definition}

\begin{observation}
\rm In this letter, a coefficient $\cal D$ is introduced to evaluate the capability of the DL model and a larger value of $\cal D$ indicates a stronger DL model. Owing that the EDNN model deployed at the edge server is generally powerful than the DNN model implemented at the $m$-th vehicle, so we have ${\cal D}_S\!>\!{\cal D}_{m}^V$, $m \in \cal M$, which leads to lower inference error rate at the server. 
\end{observation}

\theoremstyle{Remark} 
\newtheorem{remark}{{\textit{Remark}}}
\begin{remark}
\rm Although it is difficult to obtain an exact analytic formula for the mapping function in reality, the observations and conclusions in the paper do \emph{not} depend on the exact formula and would not change even the exact formula changes because the trends will remain similar: 
\textit{good data quality and stronger DL model will have less inference error} \cite{{quality_on_DL01},{quality_on_DL02}}.
\end{remark}

It is observed from Fig.~\ref{v2p_model} that a collision will occur once the vehicle and the pedestrian occupy the \textit{collision zone} at the same time. 
Once the critical event has occurred (e.g., a pedestrian who is crossing the road is detected), the vehicle then executes the recommended avoidance maneuver (e.g., by braking). Let the acceleration of the vehicle be $a_v$, which is a negative number.
Denote the total time cost for the vehicle to reach and clear the collision zone as $t^R_{v}$ and $t^C_{v}$, respectively.

From Fig.~\ref{v2p_model}, the time for the pedestrian to reach and clear the collision zone can be calculated as $t^R_{p}=\frac{D_{pz}}{V_p}$ and $t^C_{p}=\frac{D_{pz}+L_p+W_v}{V_p}$.
In order to avoid a crash in this vehicle-going-straight/pedestrian-crossing-road scenario, one of the two conditions need to be satisfied: \textit{Condition A}: the vehicle should reach the collision zone after the pedestrian clears it, i.e., $t^R_{v}\geq t^C_{p}$, or, \textit{Condition B}: the vehicle should clear the collision zone before the pedestrian reaches it, i.e., $t^C_{v}\leq t^R_{p}$. 

\begin{observation}
\rm For condition A, it may be possible to prevent an impending crash by braking if the vehicle is far enough away from the collision zone. For condition B, if the vehicle is within a certain distance of the collision zone, it may be possible to avoid a crash if the vehicle maintains the initial speed or accelerates at this distance. According to the NTSB report, the Uber accident belongs to condition A, which is exactly what happens most often but difficult to deal with.
\end{observation}

In condition A, we denote the maximum tolerable delay of ${\cal T}_m$ with pre-braking and without pre-braking as $\vartheta_m^{\rm b}$ and $\vartheta_m^{\rm ub}$, respectively. The distance between the vehicle and the collision zone is $D_{vz}$. To maximize the inference accuracy within the task latency constraint ($\vartheta_m$), the definition of braking probability is given as below.

\begin{definition}
\label{brake_p}
 \textbf{Pre-braking Probability.} \rm It is defined as the probability that the offloading vehicles perform the braking in advance when the captured images are to be identified. 
\end{definition}

The pre-braking usually happens when the inference delay ($\tau_m$) is larger than the maximum tolerable delay of the DLT, i.e., $\eta_m = {\rm Pr}\{\tau_m \geq \vartheta_m^{\rm ub}\}$. With the pre-braking, the maximum tolerable delay of ${\cal T}_m$ can be increased by $t_\Delta$, i.e., $t_\Delta=\vartheta_m^{\rm b} - \vartheta_m^{\rm ub}$, which is derived in the following Lemma and proved in Appendix~\ref{appendix_B}.
\begin{lemma} \label{L1}
\rm Additional time saved by pre-braking is given by 

\noindent $t_\Delta\!=\!\underbrace{\frac{\!-\!V^m_v\!+\!\sqrt{(V^m_v)^2\!+\!2a^m_vD_{vz}} }{a^m_v}}_{\vartheta_m^{\rm b}}\!-\! \underbrace{\left ( t_p^C \!-\! \sqrt{\frac{2 \left(V^m_v t_p^C \!-\! D_{vz} \right)}{-a^m_v}} \right )}_{\vartheta_m^{\rm ub}}$.  
\end{lemma}
 
Compared to the case without pre-braking, the task latency constraint ($\vartheta_m$) can be improved. This means that additional time can be saved for braking, i.e.,
\begin{equation}
\vartheta_m=\left ( 1-\eta_m \right ) \vartheta_m^{\rm ub} + \eta_m \vartheta_m^{\rm b}.
\end{equation}

Even though more time can be used to infer the object from the captured images, however, if the pedestrian cannot be identified correctly within the time duration of $\vartheta_m$, the crash still occurs within the collision zone. To further improve the inference accuracy, offloading the DLTs to the MES is promising because a more powerful DL model can be used to infer the pedestrian more accurately. 

\begin{definition}
\label{offloading_p}
 \textbf{Offloading Probability.} \rm For the offloading vehicle $OV_m$, the offloading probability ($\varrho_m$) is defined as the probability that the vehicle offloading their tasks to the MES. 
\end{definition}
 
Suppose that the offloading vehicle $OV_m$ can evaluate the inference error rate in near-real-time~\cite{Vehicle01}, denoted as ${\epsilon_m^L}$. If ${\epsilon_m^L}$ is above a certain threshold $\epsilon_m^{\rm th}$, then the data will be offloaded to the BS. Thus, the offloading probability of $OV_m$ equals the probability that $\epsilon_m^{L} \geq \epsilon_m^{\rm th}$, i.e., 
\begin{equation}\label{P_offloading}
\varrho_m= {\rm{Pr}}\left \{\epsilon_m^{L} \geq \epsilon_m^{\rm th} \right \} =\int_{\epsilon_m^{\rm th}}^{\infty } e^{-x}dx=e^{-\epsilon_m^{\rm th}}.
\end{equation}

\begin{observation}
\rm According to Definition~\ref{D2}, the inference error rate of the offloading vehicle $OV_m$ is  $\epsilon_m^L \!=\! g(\mathcal {Q,D}_m^V)$. It can be observed from (\ref{P_offloading}) that once the inference error rate threshold ($\epsilon_m^{\rm th}$) is given, the offloading probability $\varrho_m$ is only determined by the input data quality ($\cal Q$) because the trained DL model (${\cal D}_m^V$) is already deployed at $OV_m$.
\end{observation}

Each DLT can be characterized by a three-tuple of parameters, i.e., ${\cal T}_m(s_m, c_m, \vartheta_m)$. In particular,  $s_m$ [bits] specifies the amount of input data necessary to be processed, $c_m$ [cycles] denotes the total number of CPU cycles required to process ${\cal T}_m$, and $\vartheta_m$ [secs] denotes the maximum tolerable delay. The communication and computing models are detailed as follows.

\subsubsection{Communication Model}
Suppose that the total wireless bandwidth is $B$, which can be further divided into $M$ sub-bands for the uplink communication. Denote $p_{t}^m$ as the transmission power of $OV_m$. $h_m$ represents the channel gain between $OV_m$ and the MES. $F$ is the computation capability of the MES and $\delta^2$ denotes the background noise power. Considering that the size of the execution results at the MES is generally much smaller compared to that of input data, the downloading delay of the execution results is negligible. Therefore,  the communication delay of the uplink between $OV_m$ and the MES is calculated as 
\begin{equation}\label{uplink_delay}
C_m=s_m/\left(\frac{B}{M} {\rm log}\left (1+\frac{p_{t}^m h_m}{\delta^2}  \right ) \right).
\end{equation}

\subsubsection{Computing Model}
Let $f_m$ be the CPU computation capacity of $OV_m$, then the inference delay of $OV_m$ at local and at the MES can be respectively calculated as
\begin{equation}\label{local_delay}
\tau^{L}_m={c_m}/{f_m},
\end{equation}
\begin{equation}\label{offloading_delay}
\tau^{O}_{m}=s_m/\left(\frac{B}{M} {\rm log}\left (1+\frac{p_{t}^m h_m}{\delta^2}  \right ) \right)+c_m/{F}.
\end{equation}

It is observed from (\ref{local_delay}) and (\ref{offloading_delay}) that the inference delay at the vehicles equals to the local computing delay, i.e., ${c_m}/{f_m}$. However, the inference delay at the MES includes not only the \textit{computing delay} (i.e., $c_m/{F}$) but also the \textit{uplink communication delay} (i.e., $s_m/\left(\frac{B}{M} {\rm log}\left (1+\frac{p_{t}^m h_m}{\delta^2}  \right ) \right)$). Therefore, the inference delay at the MES is generally larger than that of vehicles due to the additional wireless communication delay, i.e., $\tau^{O}_m>\tau^{L}_m$.

Based on the critical parameters introduced previously, the overall delay and inference error rate  of $OV_m$ are obtained as
\begin{equation}\label{delay}
\tau_m=\left(1-\varrho_m\right)\tau^{L}_m+\varrho_m \tau^{O}_m=\tau^{L}_m+\varrho_m \left(\tau^{O}_m-\tau^{L}_m\right),
\end{equation}
\begin{equation}\label{inference}
{\epsilon}_m = \left(1-\varrho_m\right)\epsilon_m^{L} + \varrho_m \epsilon_m^{O}=\epsilon_m^{L}-\varrho_m\left(\epsilon_m^{L}-\epsilon_m^{O}\right).
\end{equation}

In this letter, to avoid the self-driving accidents, an inference error rate minimization problem is formulated as $\mathscr{P}$, i.e.,
\begin{equation} \label{e01}
\begin{aligned}
&{\mathscr{P}}: \underset{\{\varrho_m \in [0,1 ], \ \eta_m \in [0,1 ] \}}{\bf min}\;\; \sum_{m \in \mathcal{M} }^{}  \epsilon_m   \\
&\;\;\;\;\;\;\;\;\;\;\;\;\; \tau_m \leq \vartheta_m, \ \forall m \in \mathcal{M},
 \end{aligned}
\end{equation}
where the constraint indicates a delay bound.

\section{Optimized Offloading Framework Design}
It is observed from (\ref{inference}) that 
$\epsilon_m$ is monotonically decreasing with respect to the OV's offloading probability $\varrho_m$.
This observation makes intuitive sense as an increase of the $\varrho_m$ would lead to a lower inference error rate. It can be derived that ${\epsilon}_m$ is minimized as ${\epsilon}^{O}_m$ when ${\varrho}_m=1$.
However, due to the constraint $\tau_m \leq \vartheta_m$, the following inequality must be satisfied 
\begin{equation} \label{optimal_eq}
\tau^{L}_m+\varrho_m \left(\tau^{O}_m-\tau^{L}_m\right) \leq \left ( 1-\eta_m \right ) \vartheta_m^{\rm ub} + \eta_m \vartheta_m^{\rm b}.
\end{equation} 

By solving (\ref{optimal_eq}), we can obtain
\begin{equation} \label{optimal_eq1}
\varrho_m  \leq \frac{ \vartheta_m^{\rm ub} + \eta_m \left (\vartheta_m^{\rm b}-\vartheta_m^{\rm ub}  \right ) -\tau^{L}_m}{\tau^{O}_m-\tau^{L}_m}. 
\end{equation} 

In this letter, the DLT is considered as a Poisson arrival process, where $s_m$ and $c_m$ theoretically follow an exponential distribution~\cite{exponential}\footnote{The proposed offloading framework can be extended into other distributions with a minor modification on the calculation of $\eta_m$.}.
According to Definition~\ref{brake_p}, $\eta_m$ is given by $\eta_m= {\rm Pr}\{\tau_m \geq \vartheta_m^{\rm ub}\}=e^{-\vartheta_m^{\rm ub}}$.
Substituting $\eta_m$ into (\ref{optimal_eq1}), 
$\varrho_m  \leq \frac{ \vartheta_m^{\rm ub} + e^{-\vartheta_m^{\rm ub}} \left (\vartheta_m^{\rm b}-\vartheta_m^{\rm ub}  \right ) -\tau^{L}_m}{\tau^{O}_m-\tau^{L}_m}$ is obtained.
 
Therefore, to minimize inference error rate, the optimal offloading probability is 
\begin{equation} \label{Optimal_rou}
\varrho_m^*={\rm min} \left \{ \frac{ \vartheta_m^{\rm ub} + e^{-\vartheta_m^{\rm ub}} \left (\vartheta_m^{\rm b}-\vartheta_m^{\rm ub}  \right ) -\tau^{L}_m}{\tau^{O}_m-\tau^{L}_m}, \ 1 \right \}.
\end{equation}

Substituting (\ref{Optimal_rou}) into (\ref{delay}) and (\ref{inference}), the minimized inference error rate is obtained as ${\epsilon}_m^* =\epsilon_m^{L}-\varrho_m^*\left(\epsilon_m^{L}-\epsilon_m^{O}\right)$ and the inference delay is achieved as $\tau_m^*=\tau^{L}_m+\varrho_m^* \left(\tau^{O}_m-\tau^{L}_m\right)$.  To this end, the edge learning-aided offloading algorithm is summarized in Algorithm~\ref{alg1}. 

\begin{algorithm}[t]
\small
\caption{Edge Learning-aided Offloading}\label{alg1} 
\textbf{Input:} Captured image with data quality ($\cal Q$) and total number of vehicles ($M$) obtained from the edge server\;
\For{{\rm  vehicle} $m$}
    {Calculate the pre-braking probability ($\eta_m$)\; 
    Calculate optimal offloading probability ($\varrho_m^*$)\;
    Calculate the optimal inference error rate threshold ($\epsilon_m^{\rm th*}$)\;
    Evaluate the inference error rate, i.e., $\epsilon_m^L \!=\! g(\mathcal {Q,D}_m^V)$\; 
    \eIf {$\epsilon_m^L \geq \epsilon_m^{\rm th*}$} {Offload the DLT  to the MES\;}
    {Process the DLT locally\;}
    }
\end{algorithm}

According to (\ref{P_offloading}), the optimal inference error rate threshold is derived as $\epsilon_m^{\rm th*}={\rm ln} \left(\frac{1}{\varrho_m^*} \right)$. If the inference error rate is larger than the threshold, i.e., if $\epsilon_m \geq \epsilon_m^{\rm th*}$, then the current DLT is offloaded to the MES.
Therefore, to identify the pedestrian with maximum inference accuracy within a certain delay, the vehicle needs to perform pre-braking with the probability $\eta_m$ and offload their DLTs with the optimal offloading probability $\varrho_m^*$ derived in (\ref{Optimal_rou}). 

\begin{observation}
 \rm There exists a \textit{trade-off} between the inference error rate and inference delay, as indicated by (\ref{delay}) and (\ref{inference}). In other words, the achievement of a low inference error rate is at the expense of inference delay. Specifically, when a large portion of the data is with ``Bad" quality, we may not be able to keep the overall inference error rate (${\epsilon}_m$) small enough because the delay constraint should also be satisfied. Therefore, the accident may still occur. This suggests that the vehicle should slow down to allow ample time to ``learn'' the environment and take proper actions during challenging environments such as bad weather and poor lighting along the road.
\end{observation}
 
\section{Simulation Results and Discussions}
 In this section, we evaluate the performance of the proposed offloading framework. Since there is no existing method optimizing offloading probability to minimize inference error rate for delay sensitive self-driving services, we compare the proposed offloading framework (with legend `Proposed framework') with two benchmark schemes:
`Local inference' (inference is always done locally at the vehicle) and `MES inference' (inference is always done at the MES). The system parameters are: $f_m\!=\!1$ GHz, $p^m_t\!=\!0.3$ W, $\delta^2\!=\!7.9\times 10^{-13}$, $B\!=\!1$ MHz, $c_m\!=\!1$ Mbits, $L_p\!=\!50$ cm, $V_p\!=\!3.6$ km/h, $D_{vz}\!=\!20$ m, $D_{pz}\!=\!3$ m, $W_v\!=\!1.5$ m, $V^m_v\!=\!55$ km/h, $a^m_v\!=\!-2.5$ m/s$^2$. Besides, we assume that the mapping function is $g(\mathcal {Q,D})=\alpha \left (\frac{1-{\cal Q}}{\cal D}  \right )$, where ${\cal Q} \in [0,1]$, $\alpha \in [0,1]$ is the scaling coefficient\footnote{In this letter, the analytic formula for the mapping function is for simulation purpose only. This can help us to evaluate the impact of data quality on the inference error rate and inference delay.}.  For simplicity, we set $\alpha=1$, ${\cal D}_m^V=1$, ${\cal D}_S=5$ and  ${\cal Q}=0.05$ indicating that data quality is ``bad". For practical purpose, we could obtain numerical evaluation of the mapping function $g(\mathcal {Q, D})$ from empirical data offline and then perform table lookup and interpolation to obtain inference error rate online. Specifically, we collected the images data with different data quality levels and performed testing via the pre-trained deep learning (DL) model accordingly. The output corresponds to the accuracy (or confidence) of the DL model with considered input data with a specific quality~\cite{quality_on_DL01}. Then the methods for function approximation can be used to obtain the relationship between  $\epsilon$ and $\cal Q$ for a given DL model ($\cal D$).

\begin{figure}[t] 
\centering
\captionsetup{font={footnotesize }}
\subfigure[]{
\includegraphics[width=2.25in,height=1.85in]{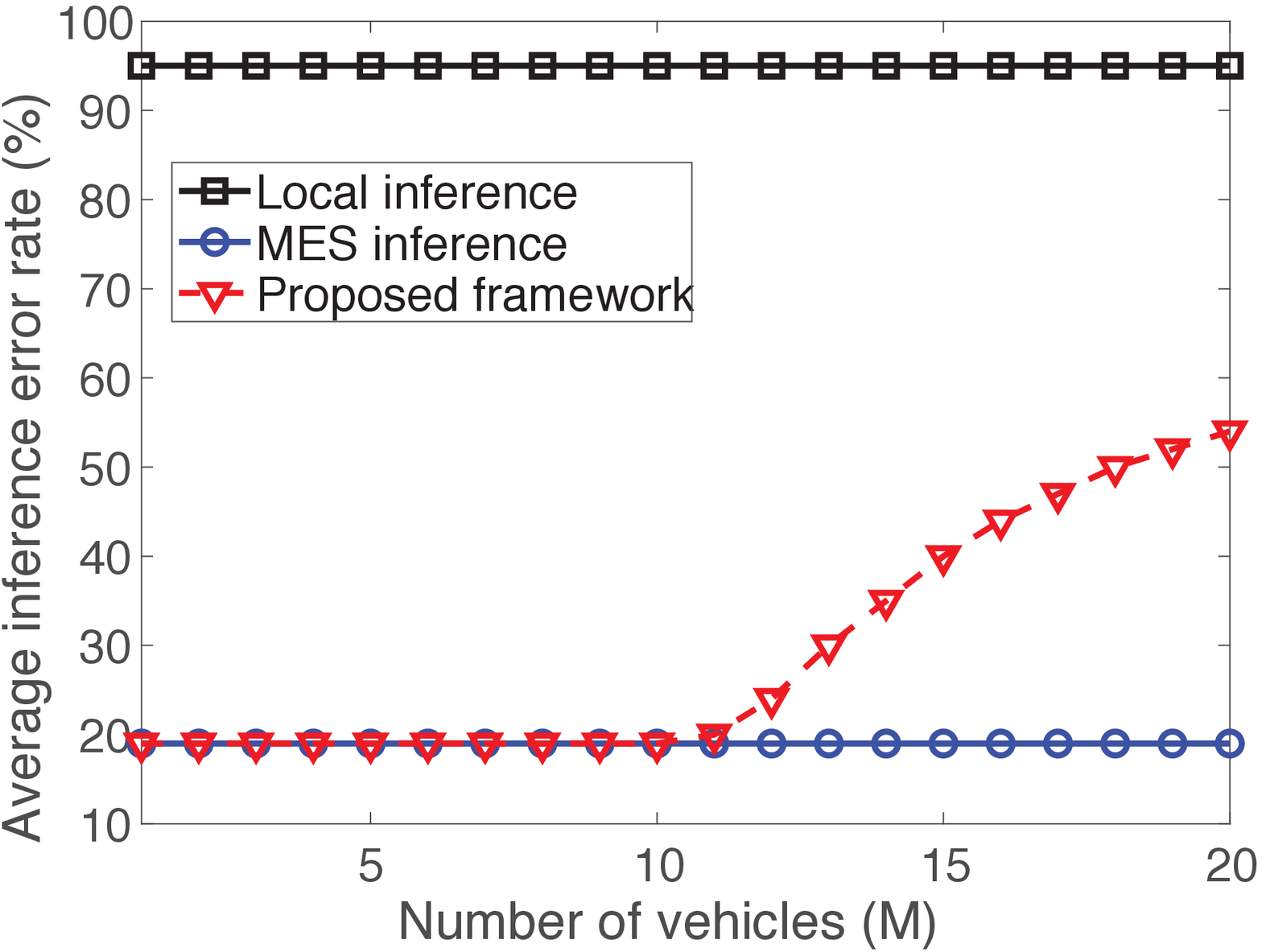}} 
\hspace{-0.00in}
\subfigure[]{
\includegraphics[width=2.25in,height=1.85in]{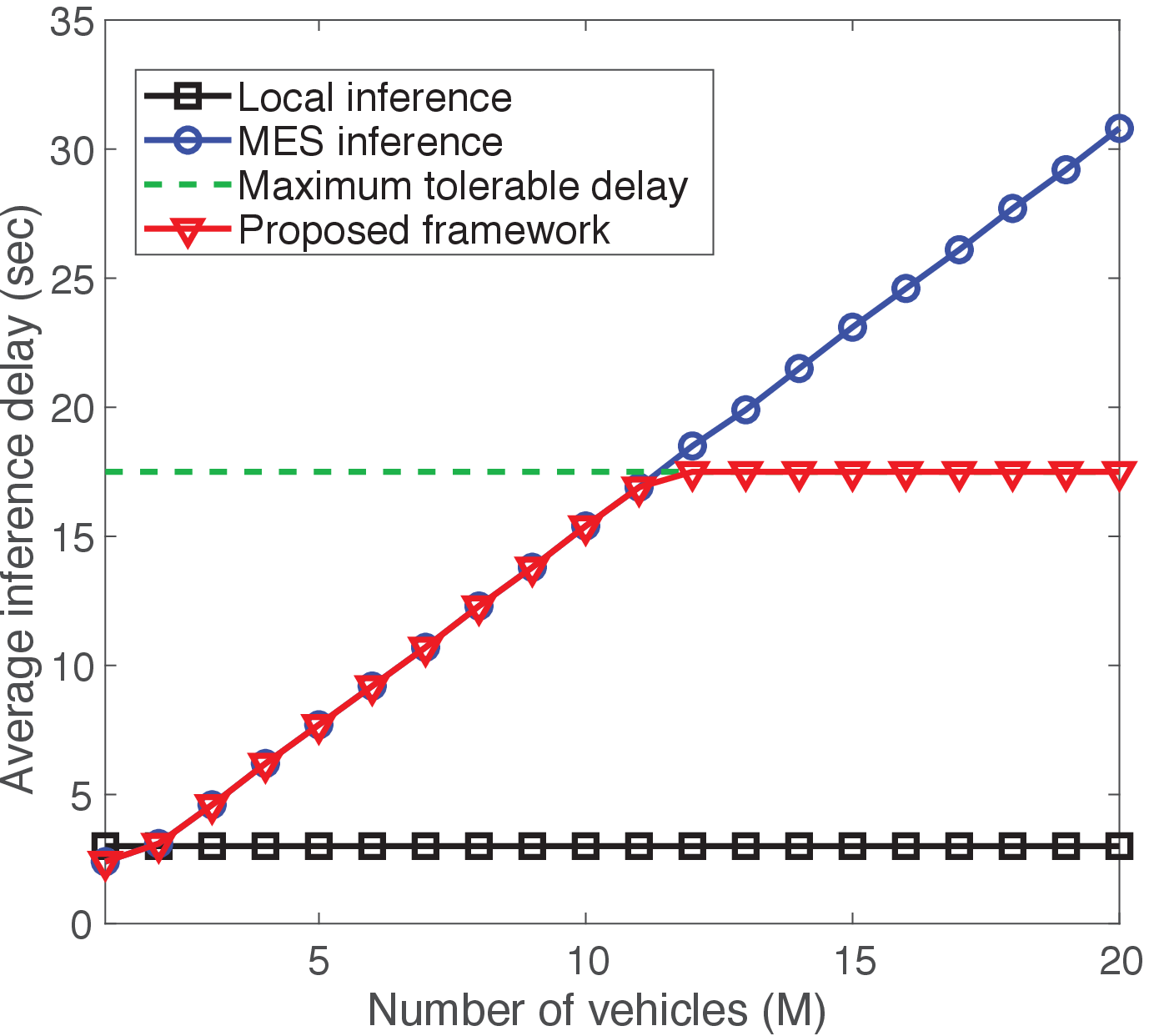}}
\caption{Average inference error rate and inference delay of the three offloading schemes v.s. number of vehicles ($M$) in (a) and (b), where $F\!=\!2$ GHz}
\label{simulation1}
\end{figure}

The average inference error rate and inference delay comparison is shown in Fig.~\ref{simulation1}(a)-(b), where $F=2$ GHz. Because the inference accuracy is mainly determined by the data quality and DL model, the inference error rate of the two benchmark schemes always keeps unchanged. As $M$ increases from $1$ to $10$, the inference accuracy and delay of the `Proposed framework' coincide with the `MES inference' scheme because the inference delay is far below the constraint given in (\ref{optimal_eq}), as also verified by Fig.~\ref{simulation1}(b). During this time, the DLTs are offloaded to MES with probability $1$. After this, the inference error rate of the `Proposed framework'  increases gradually along  $M$ while the inference delay always keeps at the threshold value. This is due to the additional delay introduced by offloading, which decreases the offloading probability. 

\begin{figure}[t] 
\centering
\captionsetup{font={footnotesize }}
\subfigure[]{
\includegraphics[width=2.25in,height=1.85in]{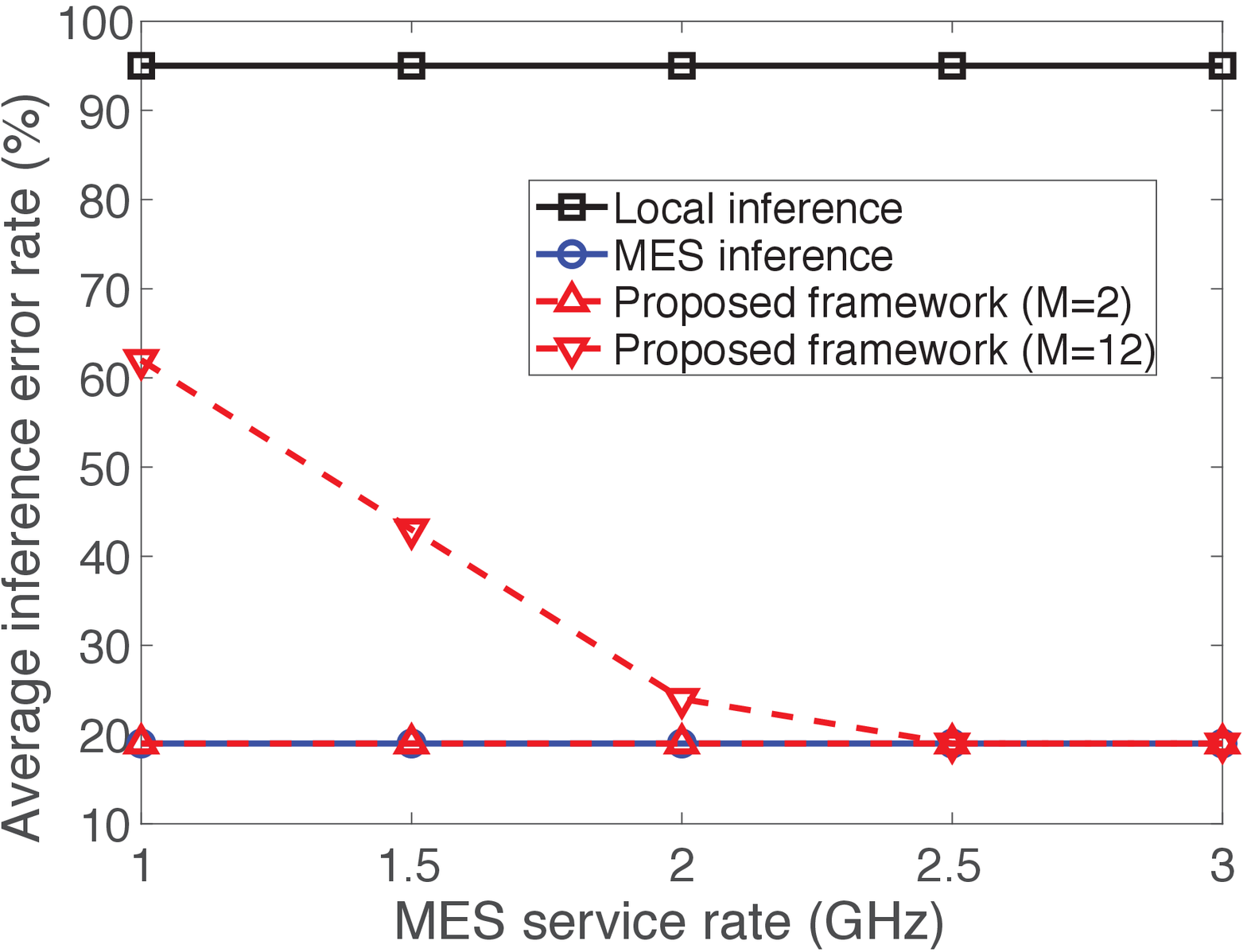}}
\vspace{-2mm}
\subfigure[]{
\includegraphics[width=2.25in,height=1.85in]{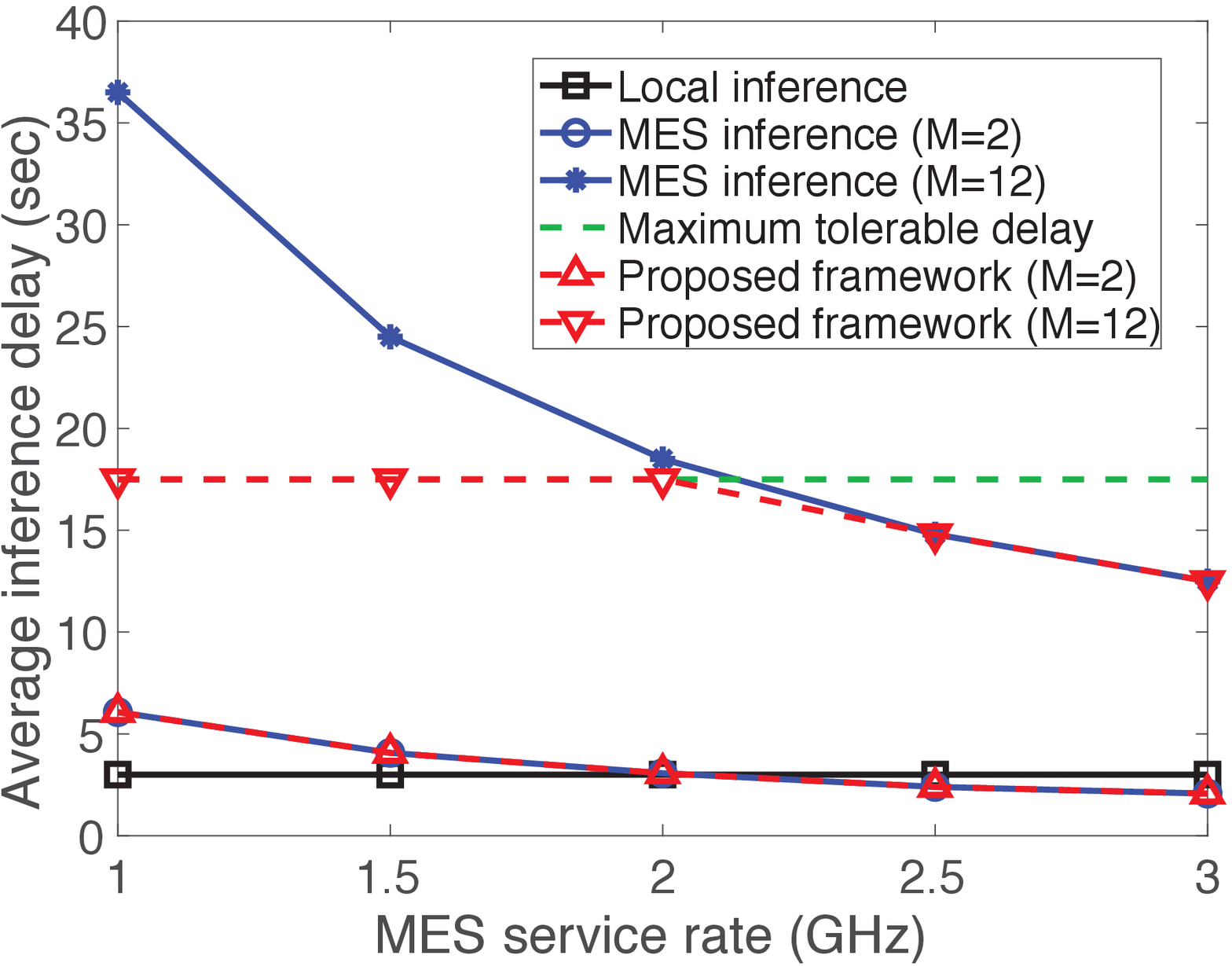}}
\caption{Average inference error rate and inference delay of the three offloading schemes v.s. MES's service rate ($F$), where $M=2$ and $12$ in (a) and (b)}
\label{simulation2}
\end{figure}

The impact of the MES service rate ($F$) on the inference error rate and delay are shown in Fig.~\ref{simulation2}(a)-(b), where $M\!=\!2$ and $12$. It is observed that the `Proposed framework' performs identically with the `MES inference' scheme when $M\!=\!2$. This is because there exist only a few number of vehicles offloading DLTs, which can be allocated with enough uplink bandwidth and computation resources at the MES. As a result, the delay constraint is always satisfied, as verified in Fig.~\ref{simulation2}(b). When $M\!=\!12$, the inference error rate of the `Proposed framework' first decreases with an increase of $F$ until $F\!=\!2.5$ GHz. The reason is that when $F$ is relatively small, the inference delay constraint can only be satisfied at the expense of the inference accuracy. 
Furthermore, the simulation results show that when $M=20$ and $F=2$ GHz, our proposed offloading framework can save around $50\%$ of inference error rate compared with the `Local inference' scheme and $45\%$ of inference delay compared with the `MES inference' scheme.

\begin{figure}[t] 
\centering
\captionsetup{font={footnotesize }}
\subfigure[]{
\includegraphics[width=2.25in,height=1.85in]{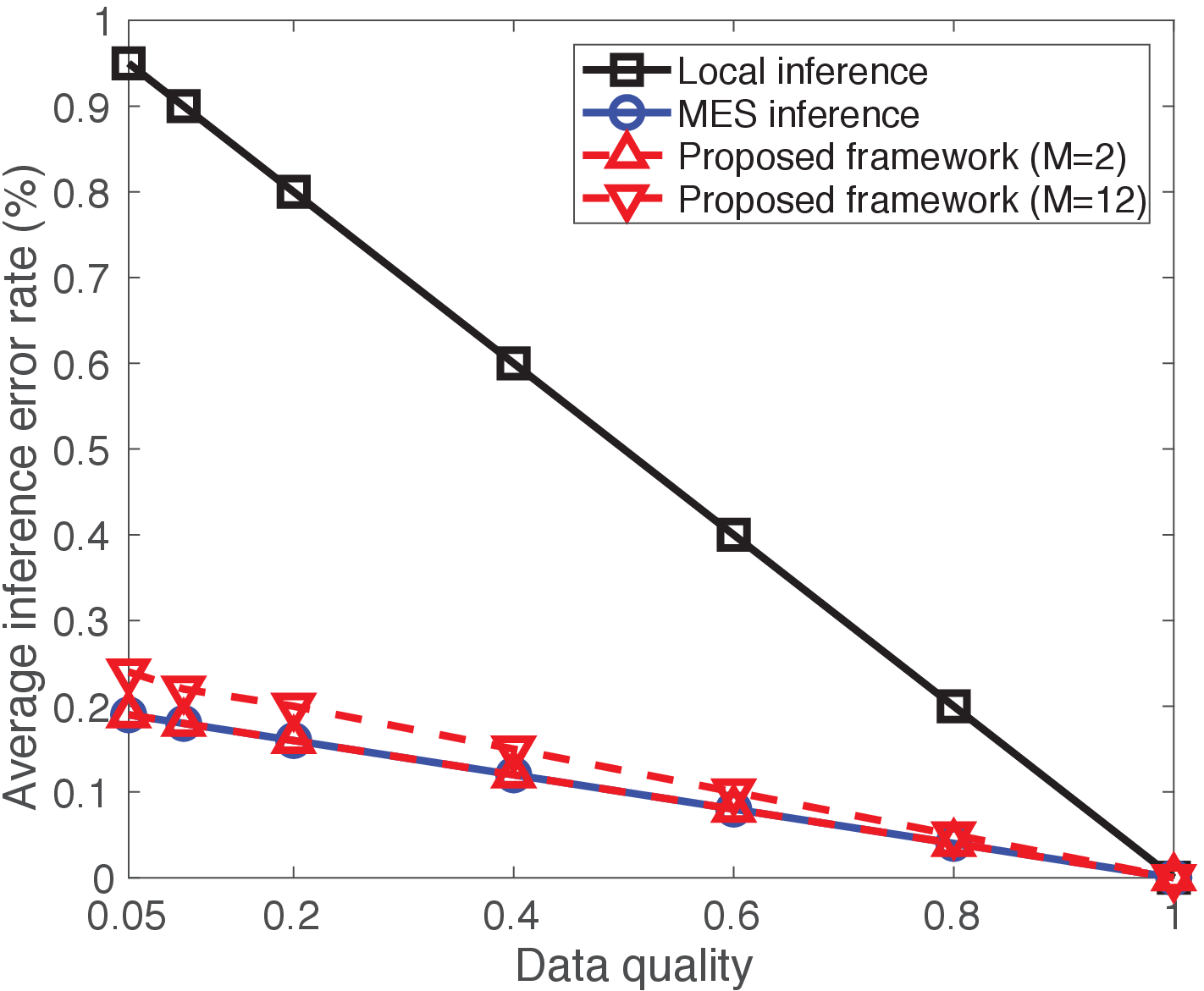}}
\subfigure[]{
\includegraphics[width=2.25in,height=1.85in]{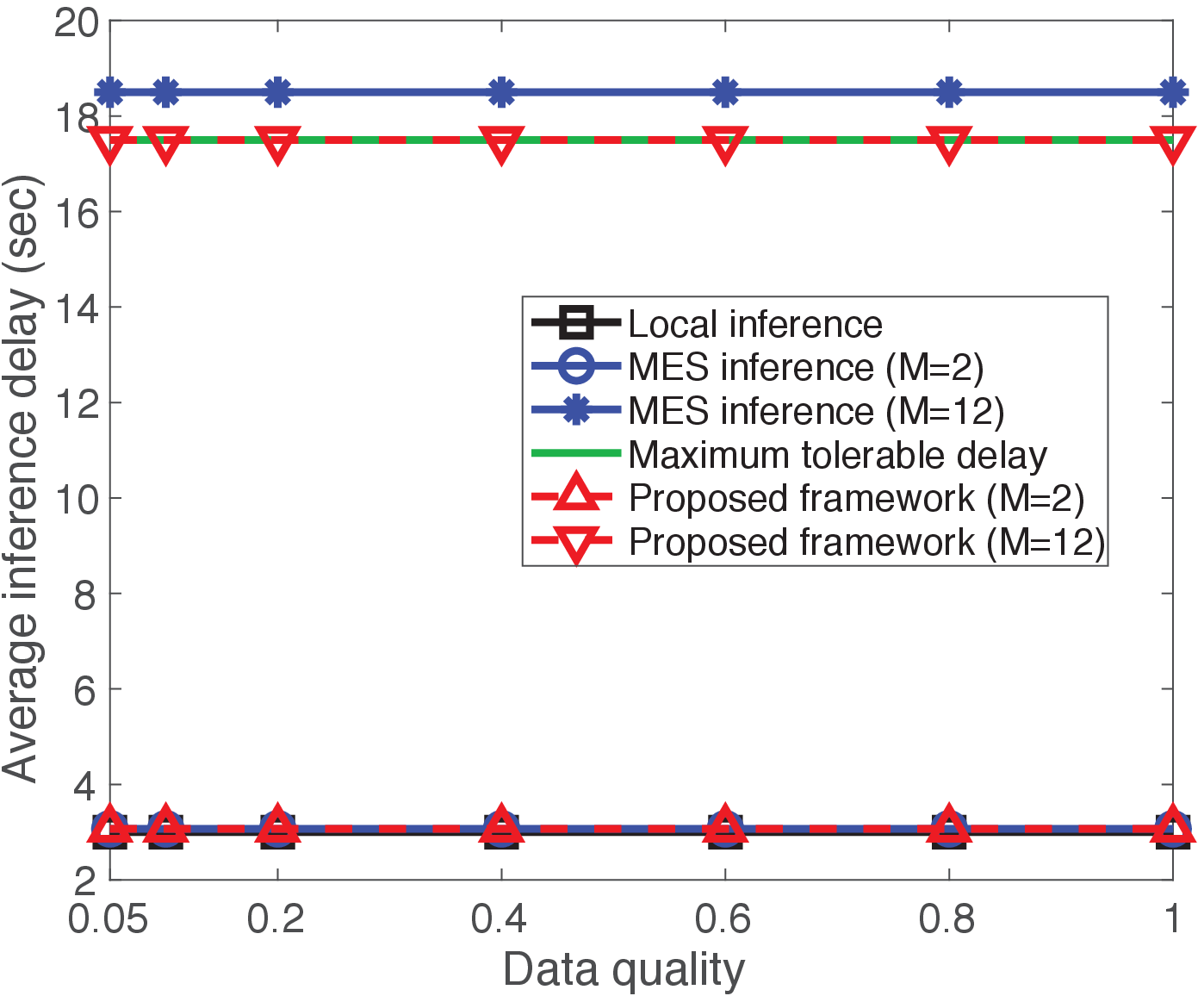}}
\caption{Average inference error rate and inference delay of the three offloading schemes v.s. data quality ($Q$), where $M=2$ and $12$ in (a) and (b)}
\label{simulation3}
\end{figure}

Fig.~\ref{simulation3}(a)-(b) demonstrate the impact of the data quality ($Q$) on the inference error rate and delay, respectively, where $F=2$ GHz, $M\!=\!2$ and $12$. From Fig.~\ref{simulation3}(a), it is observed that the inference error rate of all the schemes decreases with an increase of $Q$. This indicates that the better data quality (i.e., the larger value of $Q$) the lower inference error rate. Due to the superiority of the trained EDNN model at the MES is explored by offloading, i.e., ${\cal D}_S\!>\!{\cal D}_{m}^V$ is achieved, the inference error rate of `Local inference' is much higher than that of other three schemes, especially when data quality is `Bad' (i.e., with smaller $Q$). Besides, we can see that the inference error rate of $M=12$ is a little higher than that of $M=2$, especially when $Q$ is small. This is because a high offloading probability can be achieved when the number of vehicles is small and the data is with ``Bad" quality. As a result, the inference error rate can be decreased accordingly. The inference delay comparison is shown in Fig.~\ref{simulation3}(b), where we can observe that the delay of all schemes keeps unchanged. This is mainly because the inference delay obtained at local ($\tau_m^L$) and MES ($\tau_m^O$) are not impacted by the data quality. In addition, Fig.~\ref{simulation3}(b) shows that the inference delay of $M=12$ is much higher than that of $M=2$ but does not exceed the delay constraint. By comparing Figs.~\ref{simulation3}(a) and \ref{simulation3}(b), it is interesting to find that the inference accuracy is decreased with an increase of $Q$ while inference delay keeps unchanged. The reason behind this phenomenon is that there exists a trade-off between the inference error rate and inference delay. Therefore, the achievement of a low inference error rate is at the expense of inference delay, which is guaranteed not exceed a threshold value.

\section{Conclusion}
This letter is motivated by an autonomous vehicle accident and a novel offloading framework is proposed to minimize the inference error subject to latency constraint. The optimal offloading probability and the pre-braking probability are analyzed to evaluate the performance of inference accuracy and gain valuable design insights. The proposed offloading framework can be used to improve inference accuracy and reduce such accidents in practical self-driving scenarios.

\appendices

\section{Proof of Lemma 1}
\label{appendix_B}
The total time cost for the vehicle to reach the collision zone can be derived as 
$t^R_{v}=\frac{-V^I_v+\sqrt{(V^I_v)^2+2a_vD_{vz}}}{a_v}$. To analyze $t_\Delta$, the following two cases are considered. 
\begin{itemize}
\item {{Case 1: With pre-braking.}}  The vehicle performs pre-braking in advance to save more time to improve the inference accuracy by offloading the tasks to the MES. In this case, $\vartheta_m^{\rm b}=\frac{-V^m_v+\sqrt{(V^m_v)^2+2a^m_vD_{vz}}}{a^m_v}$ can be achieved.
\item  {{Case 2: Without pre-braking.}} In this case, the vehicle keeps at the current speed (i.e., $V^m_v$) until the pedestrian is identified. Considering that the minimum allowable time to avoid a crash by braking is $t^C_p$, then we have
\begin{equation} \label{No_b}
\vartheta_m^{\rm ub} V^m_v \!+\! V^m_v \left (t^C_p \!-\! \vartheta_m^{\rm ub} \right ) \!+\! \frac{1}{2}a^m_v \left (t^C_p \!-\! \vartheta_m^{\rm ub} \right )^2\!=\!D_{vz}.
\end{equation}

By solving (\ref{No_b}), $\vartheta_m^{\rm ub}=t_p^C \!-\! \sqrt{\frac{2 \left(V^m_v t_p^C \!-\! D_{vz} \right)}{\!-\!a^m_v}}$ is obtained. Based on $\vartheta_m^{\rm ub}$ and $\vartheta_m^{\rm b}$, $t_\Delta$ can be calculated accordingly.
\end{itemize}



\ifCLASSOPTIONcaptionsoff
  \newpage
\fi

\vspace{12pt}

%

%
%
%




\end{spacing}
\end{document}